O.A. Sychev, D.P. Mamontov

# Determining token sequence mistakes in responses to questions with open text answer

Volgograd State Technical University

oasychev@gmail.com, mamontov.dp@gmail.com

When learning grammar of the new language, a teacher should routinely check student's exercises for grammatical correctness. The paper describes a method of automatically detecting and reporting grammar mistakes, regarding an order of tokens in the response. It could report extra tokens, missing tokens and misplaced tokens. The method is useful when teaching language, where order of tokens is important, which includes most formal languages and some natural ones (like English). The method was implemented in a question type plug-in CorrectWriting for the widely used learning manage system Moodle.

## 1. Introduction

There was increased interest to automatic determining of mistakes and grading of student's responses to the question with open answer (usually text) in the latest years. However, most of such works was devoted to the grading of natural language answers [1], which is useful in teaching many courses, requiring thorough answers in the student's native language.

Far less solutions aimed to the learning the grammar of the language, while this task is important in courses where student should learn a new language – either natural (usually foreign) or formal (for example, programming language in the introductory programming courses) one. There is significant difficulty in the learning grammar rules of the new language [2]. Having a question that can automatically report some common mistakes in grammatical constructs of the learned language will allow students to train their ability to write correctly in the new language and ease the burden of the routine checking of grammar exercises for the teachers.

A lot of solutions allow computing the closeness of the student's response to the teacher's answer. But only small number of them, based on approaches like checking matches with regular expressions or using a subset of the language compiler, gives possibility to determine mistakes or give hints. Looking for a good solution we must keep in mind the kind of hints this method could give to the student.

A good way to represent student's mistakes is to see them as editing operations, which should be done to convert student's response to the correct answer. In that case calculating editing distance allows us to find a minimal set of such operations, while the kind of distance will determine which kinds of mistakes we could report. One group of mistakes commonly made when learning grammar is sequence mistakes. Teaching grammar you often find yourself teaching to write correct words in the correct order. Sequence of tokens (lexeme) important in most formal language and several natural ones (like English language, while Slavic languages often allow almost any possible order of words in a sentence). Most common sequence errors are misplacement, insertion and deletion of characters of tokens.

The level of analysis is also important: there are well-known algorithms used to find mistakes on the character level, developed to find typos in the text processing software. However, the grammar mistakes are done either at token level or as specific mistakes inside a token (like wrong suffixes to the words). Character-level calculation of editing distances is useless to find token-level errors, and they will misrepresent grammar errors inside a token as typo.

## 2. Existing methods of automatic determining of mistakes

One important group of questions, devoted to the open answer processing, consists of the questions, using some sort of templates to allow teacher to describe all possible correct answers. There are several such question types for the widely used learning manage system (LMS) Moodle: PMatch (developed at the Open University), RegExp (developed by Joseph Rezeau, University Rennes 2) and Preg (developed at the Volgograd State Technical University).

PMatch question type [1] uses custom template language, adapted to natural language processing. It was created to automatically verify short free natural language answers and allows finding response correctness considering possible typos, synonym use, flexible word order etc. It shows a fairly good accuracy in determining natural language answer correctness, but when response contains a recognized grammar error, it tends to ignore it, rather than show to the student. It aimed to grade semantics, not the syntax.

RegExp[3] and Preg[2] question types use a well-known regular expressions (heavily limited in case of RegExp) to define templates of the correct answers. They both were originally developed for language learning: the English as foreign language (RegExp) and C++ programming language (Preg). They both can detect a correct start of the response and the place, where it biased from the correct answer, showing the student where his first mistake was made. They could hint to the student next correct character or next word (next token in case of Preg, allowing teacher to select either English or C++ language tokenizing) for a grade penalty. A student could use some hint to recover from error, but it often lead to guesswork from the student instead of learning the grammar rules.

More important is the ability of both question types to give mistake messages. The RegExp plug-in has a sophisticated system of looking for missing words and could report them to the student. Preg question type allows similar detection of missing tokens using specially written regular expressions, it also allows a crude detection of misplaced tokens (only in cases with small number of mistakes and tokens in question should be used in the answer only once). Such mistakes' reporting is much more useful in learning grammar, but these question types have several major drawbacks. Using mistakes messages require a heavy authoring work from the teacher, who should manually write the template and the message for each mistaken word (token) in each question. Misplaced token mistakes could be determined correctly by Preg only in absence of other mistakes in enclosing tokens.

A much better approach to looking for grammar errors was used in JITS system for learning Java programming language, developed by the E.R. Sykes and F.F. Franek at McMaster University [4]. It is based on the use of Java compiler, but modifies it to detect grammar mistakes. It could find such mistakes as replacement of token to another token, inserting an extra token, removing tokens, and exchange of adjacent tokens. Teacher has the ability to set an explicit answer to be given by the student, to make mistake reporting more correct. It is used to detect typos and lexical mistakes using standard per-character editing distances.

Grammar mistakes search was implemented by tracking all possible transformations of the response (and delete tokens, token insertion, substitution and exchange of tokens in places). This is quite time-consuming method, especially in cases with several mistakes in one response. The module looking for grammatical mistakes also not consider teacher-entered correct answer if available, it's just try to make student's answer comply with Java language grammar. Another drawback of JITS is hidden in the kind of mistakes it report. While such mistake type as replacement is very useful on per-character basis to find typos, replacement of whole token occurs very rare and usually can be considered as semantic, not grammar error. And when it comes to such important mistakes as token misplacement, JITS could only detect exchange places of adjacent tokens, other misplacements will be reported as a pair of insertion and deletion of the same token, which may confuse students. And, last but not least, converting JITS approach to the other target languages (formal or natural) requires a good deal of work.

### 3. Proposed method

Most common editing distances was developed to look for typos in the text editing software (see Levenshtein and Damerau-Levenshtein distances for example) failed to report misplacement kind of mistakes[5]. In this paper we propose another method, involving calculating of longest common subsequence (LCS) [6] between response and correct answer, presented as a sequence of tokens. Such subsequence could be considered a valid part of the student's response.

When LCS is lesser than correct answer, mistakes occur. A token, presented in both answer and response, but not in LCS, can be considered misplaced. A token, mentioned only in correct answer is missing in response, while the token mentioned only in response is extraneous and should be deleted. Such analysis should take into account the number of equal tokens, e.g. if the correct answer contains three copy of some token, the LCS – only one and the response – two, than one copy of such token was placed correctly, another one misplaced and third is missing in response.

Consider a question, requiring a student to write a function header "*void function(int abc, int def)*". Suppose the student's response is "*function int abc, int def, void*". The method, proposed in this paper, yield 4 mistake messages:

1) "void" is misplaced;
2) there is extra ",";
3) "(" is missing;
4) ")" is missing;

**4. Discussion**

JITS system yield 5 mistake messages in this case, splitting first mistake with "void" in the extraneous and missing token mistakes. Another advantage of proposed method of LCS analysis over the one used in JITS is simplicity of tuning it on any language with strict token order – the only language-dependent part in it is tokenizer, which could be written relatively easy using modern tokenizer-generator software like JFlex. Small deviation of strict order, allowed by the language, could be handled using several teacher-entered correct answers.

Preg and RegExp question types will show less mistakes messages than proposed method: Preg omit extra "," and RegExp omit both extra "," and misplaced "void". Authoring of question using proposed method will require less work, since teacher doesn't need to write regular expression and full messages for each token that could be missing or misplaced.

The proposed method was implemented in the CorrectWriting question type for Moodle LMS and will be available under GPL in Plugins section of the Moodle main site.

One practical consideration when teaching grammar with such question type is that you as teacher often don't want to disclose actual token text in mistake message to the student. The grammatical role of "void" keyword in example above could be described as "return value type". When supplied with message like "void is misplaced" students often tend to rely on guesswork, trying to find a place for such word. Showing "return value type is misplaced" message instead will make students think "What is return value type in this case and where it should be placed?" which is more useful in learning grammar. CorrectWriting question type deal with this, allowing the teacher to enter grammatical descriptions for each token in the correct answer, like:

1) void - return value type;

2) function - function name;

3) ( - opening bracket for arguments list;

4) int - first argument type;

5) abc - first argument name;

6) , - argument list separator;

7) char - second argument type;

8) def - second argument name;

9) ) - closing bracket for arguments list.

Such descriptions (when given) used in misplaced and missing tokens mistake messages instead of token text; obviously extra token message can't make use of grammatical description, since teacher can't predict what extra tokens student will write.

Volgograd State Technical University plans to use developed CorrectWriting question in the teaching "Basic Programming" course on computer science faculty. It is possible to use it for teaching English as foreign language too. Further developed of the question type, extending range of reported mistakes, is also possible. It will benefit from the subsystem of comparing single tokens, which could report typos and mistakes in the tokens (like incorrect word form used).